\documentclass[10pt]{article}
\ifdefined\ANON
\usepackage{tmlr}
\else
\usepackage[preprint]{tmlr}
\fi
\usepackage{amsmath,amsfonts,bm}

\def\eqref#1{equation~\ref{#1}}
\def\1{\bm{1}}

\DeclareMathAlphabet{\mathsfit}{\encodingdefault}{\sfdefault}{m}{sl}
\SetMathAlphabet{\mathsfit}{bold}{\encodingdefault}{\sfdefault}{bx}{n}

\usepackage{amsmath,amssymb}
\usepackage{booktabs}
\usepackage{graphicx}
\usepackage{microtype}
\usepackage{hyperref}
\hypersetup{hidelinks}
\usepackage{url}
\title{Iterate or Widen? When Test-Time Refinement Helps LiDAR Scene Completion\\{\large A Controlled Study of Evidence Geometry, Training Coverage, and Compute}\thanks{OpenAI ChatGPT/Codex assisted with code development, experiment orchestration, data analysis, and language editing. The authors take full responsibility for the content of the manuscript.}}
\ifdefined\ANON
\author{\name Anonymous authors}
\else
\author{
\name Shijie Hao \\
\addr Phillips Exeter Academy
\AND
\name Weining Zhang \\
\addr Cheung Kong Graduate School of Business
}
\fi
\def\month{MM}
\def\year{YYYY}
\def\openreview{\url{https://openreview.net/forum?id=XXXX}}
\begin{document}
\maketitle
\begin{abstract}
Should a completion model spend extra test-time compute by iterating, or spend a similar parameter budget on a wider one-shot predictor? The answer is easily confounded by denoising curricula, corruption augmentation, capacity, and unpaired evaluation. We study this question in LiDAR semantic scene completion by comparing a one-shot predictor, a parameter-matched wider predictor, and a weight-tied multigrid refiner initialized from the same frozen predictor. The protocol separates coherent region removal, independent thinning, range-dependent attenuation, and additive clutter while preserving exact scene-condition pairing. Across five training seeds and 815 SemanticKITTI sequence-08 frames, the full iterative system improves mIoU over the wide control by 0.911 points under contiguous angular removal, with a 95\% moving-block bootstrap interval of [0.804, 1.040] that clears a predeclared 0.5-point practical margin. Under independent 75\% thinning, iteration adds only 0.300 points [0.166, 0.436], whereas observation-family augmentation adds 5.975 points [5.662, 6.140]. Neither intervention repairs additive clutter. The iterative system also costs 10.74 ms and 0.75 GiB per frame, versus 6.25 ms and 0.23 GiB for the wide control. These results establish a geometry-conditioned empirical boundary rather than a universal advantage: coherent gaps can justify fixed-depth refinement, broadly thinned evidence is addressed more effectively by training coverage, and spurious evidence requires a different robustness mechanism.
\end{abstract}

\section{Introduction}

A recurrent or iterative predictor can reuse a small set of parameters while spending additional computation on a difficult input. A wider feed-forward model instead spends its budget in one pass. This iterate-or-widen choice now appears across reasoning systems: hierarchical and tiny recursive models refine explicit answers over cycles, looped transformers reuse layers as sequential depth, and recurrent-depth language models scale test-time computation without proportional parameter growth \citep{wang2025hrm,jolicoeur2025trm,giannou2023looped,saunshi2025reasoning,yang2024looped,geiping2025scaling,zhu2025looped}. The same choice is consequential in semantic scene completion (SSC), where a sparse LiDAR sweep must support dense voxel prediction under a strict latency budget.

The useful strategy may depend on evidence geometry. Independent thinning leaves nearby support scattered throughout the scene, while a coherent missing sector removes all local observations over a connected region; additive clutter presents a different challenge by inserting false support. Iteration is a plausible way to propagate corrections across a coherent gap, but the same claim need not hold when the remaining evidence is locally dense or actively misleading. This motivating distinction is weaker than a causal ``evidence-distance law'': establishing such a law would require direct, corrected manipulations of distance and topology at fixed removed area.

The empirical question is also deceptively confounded. Iterative models are commonly trained with deep supervision, noise injection, input recall, and denoising objectives that are absent from their one-shot baselines. A gain under corrupted input may then come from the observation model seen during training rather than from test-time iteration. Adding a refiner also adds parameters unless the one-shot comparison is widened. Finally, SSC frames are consecutive and highly correlated, while random corruptions can differ across models unless the masks are frozen. Robustness work in 2-D vision has repeatedly shown why these controls matter: apparent architectural gains often shrink after accounting for clean accuracy, simple augmentation, and the match between training and test corruptions \citep{hendrycks2019benchmarking,taori2020measuring,rusak2020simple,mintun2021interaction}.

We isolate these factors on the SemanticKITTI SSC task \citep{behley2019semantickitti}. A compact one-shot model supplies an explicit semantic-logit belief. A 0.53M-parameter weight-tied multigrid module repeatedly updates that belief. Its strongest one-shot control is widened to match the full iterative system within 0.51\% of total parameters. Clean and augmented versions share the same label mapping, input representation, optimization budget, and corruption implementation. Every checkpoint is evaluated on identical scene-condition pairs, including exact corruption-mask hashes.

The study makes four contributions:

\begin{enumerate}
\item It provides a matched iterate-or-widen design that controls
training-distribution coverage and parameter count, then tests how the full iterative system depends on its correction curriculum and coarse scale.

\item It establishes an empirical boundary across coherent removal, independent
thinning, range-dependent attenuation, and additive clutter, while avoiding a causal claim about evidence distance that the design does not identify.

\item It evaluates iterative-system effects with paired hierarchical
moving-block bootstrap intervals over training seeds and temporally ordered frames, together with a predeclared 0.5-mIoU practical margin.

\item It audits official label semantics and sensor geometry, tests whether the
observation-coverage finding transfers to the recognized LMSCNet-SS topology, and reports accuracy jointly with latency, memory, FLOPs, and parameters.

\end{enumerate}

The intended inference is deliberately conditional rather than universal. The design asks whether extra sequential depth corrects errors that a matched one-shot model leaves unresolved, and whether robustness under a shifted observation model is instead explained by exposure to that shift during training.

\section{Related Work}

\subsection{Weight-tied iteration and test-time compute}

Weight tying converts repeated application of one update rule into sequential depth without proportional parameter growth. Hierarchical Reasoning Models couple fast and slow recurrent modules, whereas Tiny Recursion Models emphasize a smaller network that repeatedly revises an explicit answer \citep{wang2025hrm,jolicoeur2025trm}. Looped-transformer analyses characterize tasks for which reused depth can substitute for a deeper feed-forward computation \citep{giannou2023looped,saunshi2025reasoning,yang2024looped}, and recurrent-depth language models report continued test-time scaling on reasoning tasks \citep{geiping2025scaling,zhu2025looped}. Deep-equilibrium models formalize the limiting computation as a fixed point \citep{bai2019deep,bai2021stabilizing}, while path-independence and algorithmic-extrapolation studies show that the training recipe determines whether recurrent computation generalizes beyond its trained horizon \citep{anil2022path,bansal2022endtoend,schwarzschild2021can}. These results motivate iteration, but they do not imply that extra steps are always helpful; overthinking near or beyond the trained depth remains an empirical possibility.

\subsection{LiDAR semantic scene completion}

SSCNet formulated semantic scene completion as the joint prediction of occupancy and semantic labels beyond a visible depth surface \citep{song2017semantic}. SemanticKITTI then established the large-scale outdoor LiDAR benchmark used here \citep{behley2019semantickitti}. LMSCNet demonstrated that a 2-D multiscale backbone with 3-D segmentation heads can offer a favorable accuracy--efficiency trade-off \citep{roldao2020lmscnet}. Later systems improve geometric context, completion-specific transfer, robustness, or long-range modeling through sparse convolutions, completion priors, recurrent scene reasoning, state-space models, and voxel detection objectives \citep{yan2021sparse,mei2023sscrs,xia2023scpnet,zhang2023occformer,li2024occmamba,li2025voxdet}. Their principal question is benchmark accuracy. Ours is narrower: we use compact models and strict controls to ask whether repeated inference itself contributes to completion and robustness. LMSCNet-SS serves as an external anchor for the observation-coverage result and corruption ranking, not as a replication of the custom iterative-system contrast or as a state-of-the-art claim.

\subsection{Iterative refinement in dense vision and SSC}

Dense vision provides several templates for iterative prediction. RAFT and its descendants maintain an explicit field, reread fixed evidence, and predict residual updates \citep{teed2020raft,teed2021droid,wang2024searaft}; iterative error feedback and uncertainty-gated partial refinement likewise allocate multiple corrections to structured outputs \citep{carreira2016human,kirillov2020pointrend}. DiffusionDet shows that a denoising-trained update can benefit from iteration even when repeatedly applying a discriminative head does not \citep{chen2023diffusiondet}. These designs motivate our explicit logit belief, input recall, residual update, and correction curriculum.

Iteration is also no longer foreign to scene completion, but existing systems use it for different information sources. TALoS updates a pretrained LiDAR SSC model at test time using line-of-sight supervision assembled from observations at other moments \citep{jang2024talos}. MonoMRN follows a monocular coarse prediction with a masked sparse GRU and reports robustness under input disturbances \citep{wang2025monocular}. DiffSSC applies a learned diffusion process in point and semantic space \citep{cao2024diffssc}, OccGen uses iterative generation for 3-D occupancy \citep{wang2024occgen}, and ESSC-RM places a feed-forward 3-D U-Net refinement module after a coarse camera-based prediction \citep{zhang2025esscrm}. These studies establish the relevance of refinement. They do not isolate, for a fixed single LiDAR scan, whether reusing one update rule is preferable to an equally trained one-shot network with comparable parameters, or whether a robustness gain is explained by corruption exposure during training. Our model performs no test-time optimization and consumes no future frame.

\subsection{Robustness and attribution}

ImageNet-C established evaluation on fixed corruption families \citep{hendrycks2019benchmarking}. Subsequent work showed that much of apparent robustness under distribution shift can be explained by clean accuracy or training data, that simple noise augmentation is a strong baseline, and that augmentation transfers in proportion to its resemblance to the test corruption \citep{taori2020measuring,rusak2020simple,mintun2021interaction}. LiDAR and bird's-eye-view benchmarks extend the lesson to 3-D sensing: weather, external interference, sensor failure, augmentation, and representation need not share one solution \citep{kong2023robo3d,xie2023robobev}. A controlled SSC study adds two requirements. The corruption must define a specific observation shift, and all models must receive exactly the same realization. Missing and spurious returns are not interchangeable. We consequently separate uniform thinning, contiguous field-of-view loss, range-dependent attenuation, and near-range clutter, while matching training-family exposure before attributing a residual gain to the iterative system.

\subsection{Hole geometry and adaptive computation}

Image inpainting has long treated hole geometry as a first-order difficulty variable: partial convolutions, large-mask models, and diffusion-based resampling target regions whose interiors lack nearby support \citep{liu2018partial,suvorov2022lama,lugmayr2022repaint}. This literature motivates asking whether coherent missing regions create more demand for repeated spatial correction than dispersed thinning. Our protocol-v2 comparison establishes a geometry-conditioned boundary between registered corruption families, but it does not causally identify nearest-evidence distance. Earlier distance-stratified and fixed-area multi-wedge analyses are therefore retained only as hypothesis-forming material in Appendix~\ref{app:exploratory}.

When test-time computation is useful, it can be allocated by global scaling, learned per-token depth, early exits, or adaptive halting \citep{snell2024scaling,raposo2024mixture,huang2018multi,graves2016adaptive,banino2021pondernet}. Our setting highlights a simpler deployment question: whether to use a fixed-depth refiner at all. The answer must combine evidence geometry with measured latency and memory, because parameter matching does not imply compute matching.

\section{Research Question and Hypotheses}

Let \(o\) be a partial LiDAR observation, \(y\) the complete semantic grid, and \(L_0=f_\phi(o)\) the logits produced by a one-shot network. The iterative system applies a weight-tied update

\[
L_{j+1}=L_j+\Delta_\theta(L_j,o,s_j), \qquad
s_{j+1}=U_\theta(s_j,L_j,o),
\]

for \(j=0,\ldots,J-1\). The comparison is not whether the iterative model has more capacity, but whether repeated application of \(\Delta_\theta\) improves the prediction relative to a parameter-matched one-shot function.

We organize the analysis around three hypotheses.

\textbf{H1: evidence geometry and sequential-depth demand.} Iteration should help most when completing a region requires spatial correction beyond what the matched one-shot computation resolves. A contiguous missing region is a stronger candidate than independent thinning because it removes all local evidence over a coherent area. This hypothesis predicts a difference between registered corruption geometries; it does not by itself identify nearest-evidence distance as the causal variable.

\textbf{H2: stability at additional depth.} If the learned update is locally contractive around a useful fixed point, accuracy should improve or saturate and update magnitudes should shrink with depth. If not, accuracy should peak near the trained depth and later updates should continue changing predictions.

\textbf{H3: observation-model-family coverage.} For a shift \(\tilde o\sim q(\tilde o\mid o)\), robustness should depend strongly on whether training exposes the model to the same corruption mechanism. Our removal tests are only partially covered: training includes the same sector and thinning families but the registered 40\% and 75\% points exceed the training maxima of 30\% and 50\%. Clutter is not present in training. The comparison therefore distinguishes family-level exposure and severity extrapolation from a wholly uncovered additive shift. At matched augmentation, a residual refiner--wide difference estimates the effect of the full iterative system rather than architecture alone. The no-denoising and two-level controls then test whether that pattern depends on the correction curriculum or the coarsest recurrent scale. A gain that vanishes after input augmentation is instead more consistent with an observation-model-coverage explanation.

These are testable organizing hypotheses, not theorems about learned networks.

\section{Method}

\subsection{Dataset and official label semantics}

We use the SemanticKITTI SSC voxel task \citep{behley2019semantickitti}: ten training sequences with 3,834 frames and validation sequence 08 with 815 consecutive frames. Each input is a binary occupancy grid of \(256\times256\times32\) voxels at 0.2 m resolution; the target contains free space and 19 semantic classes. The physical axes are forward \(x\in[0,51.2]\) m, lateral \(y\in[-25.6,25.6]\) m, and vertical coordinate z.

Protocol v2 follows the official completion remapping: raw label zero is free, whereas every nonzero raw label mapped to learning id zero is ignored (255). Unknown raw uint16 ids are ignored. The invalid mask is applied after remapping. We additionally load the official first-view \texttt{.occluded} mask distributed with the voxel labels \citep{semantickitti_format}. Because the raw mask includes directly observed occupied surfaces, the stable analysis uses three disjoint regions defined before corruption: (i) originally observed surface voxels, (ii) voxels marked by the official mask after removing those surfaces, and (iii) all other originally unobserved valid voxels. This definition prevents overlap between region scores and keeps the evaluated regions fixed as corruption severity increases. For mechanism analysis, we separately record original returns retained or removed by the corruption and locations receiving artificial clutter.

\subsection{Models and matching}

\textbf{One-shot base.} The input representation contains occupancy and an occupied-height feature. Height bins are folded into channels, followed by a 2-D column encoder, five residual blocks, and a channel-to-height semantic decoder. The standard width-192 model contains 3,683,296 parameters.

\textbf{Wide one-shot control.} Increasing the trunk width to 206 yields 4,195,234 parameters. This is the primary capacity control.

\textbf{Weight-tied multigrid refiner.} The width-192 base is frozen and initializes the latent semantic logits \(L_0\). At every step, the refiner re-encodes the raw input and combines four signals: the encoded evidence, a projection of the current logits, an innovation signal measuring observed occupancy unexplained by the current belief, and predictive entropy. A shared ConvGRU smoother operates at fine, middle, and coarse scales in a V-cycle. The downward path aggregates context, the coarse state spans the scene at reduced resolution, and the upward path merges this context back into fine-scale residual corrections. The final head predicts \(\Delta L\) and is zero-initialized, so the untrained refiner exactly reproduces the base. The refiner contains 533,392 trainable parameters; the full system contains 4,216,688, only 0.51\% more than the wide one-shot control. The same refiner weights are reused for all \(J\) steps.

\begin{figure}[t]
\centering
\includegraphics[width=\linewidth]{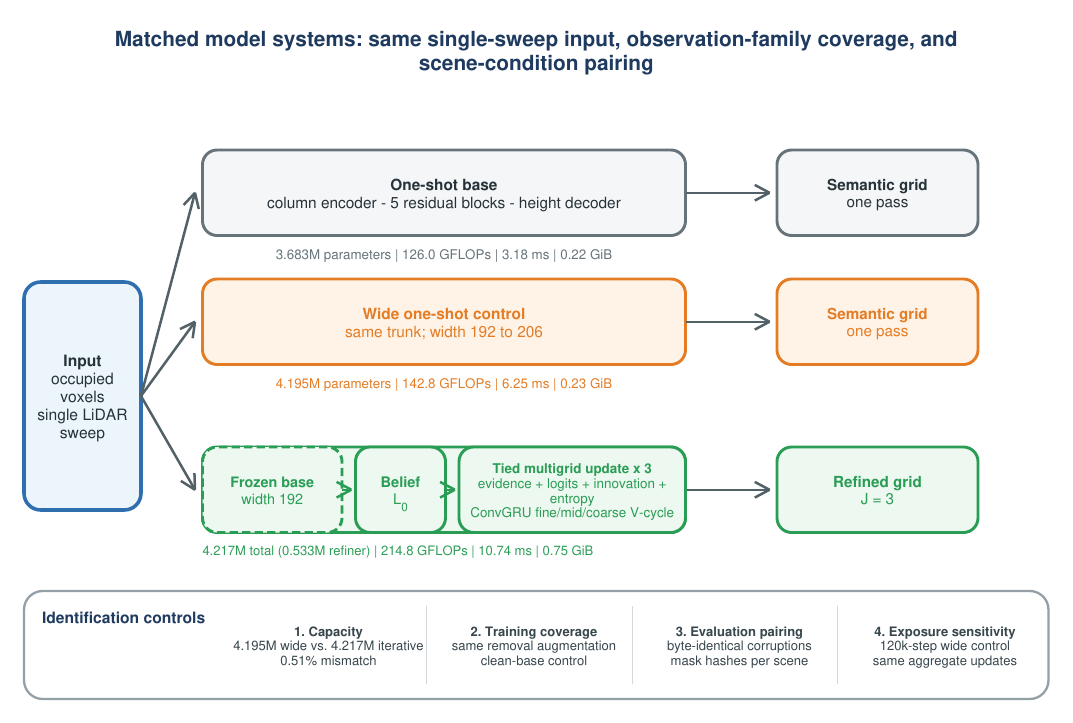}
\caption{Matched iterate-or-widen design. The one-shot and wide controls spend their parameter budget in one pass; the refiner freezes the standard base and reuses a 0.533M-parameter multigrid update for three steps. Parameter count, observation-family coverage, corruption realization, and aggregate training exposure are controlled separately. Latency, FLOPs, and memory are measured rather than inferred from parameter count.}
\label{fig:setup}
\end{figure}

\textbf{External observation-model anchor.} We reproduce the published LMSCNet-SS topology \citep{roldao2020lmscnet} from frozen upstream commit \texttt{ea1b42d...} under Apache-2.0. The adapter only permutes tensor axes and exposes the common evaluation interface. A parity test over all 80 state tensors produces bitwise-identical FP32 output to the upstream class. Clean and removal-augmented variants are trained for 80 epochs over three seeds using the author code's Conv2d initialization, random lateral flips, Adam optimizer, class weights, and per-epoch 0.98 learning-rate decay.

\subsection{Training arms}

The primary factorial comparison contains:

\begin{itemize}
\item \textbf{Clean one-shot} (\texttt{base\_clean\_v2}): the standard one-shot model trained
without removal augmentation;

\item \textbf{Augmented one-shot} (\texttt{base\_aug\_v2}): the same model trained with removal
augmentation;

\item \textbf{Wide augmented one-shot} (\texttt{wide\_aug\_v2}): the parameter-matched wider
model trained with the same augmentation;

\item \textbf{Clean-base refiner} (\texttt{v2\_cleanbase\_aug\_v2}): the augmented refiner
initialized by the clean one-shot base;

\item \textbf{Augmented-base refiner} (\texttt{v2\_augbase\_aug\_v2}): the augmented refiner
initialized by the augmented one-shot base.

\end{itemize}

The last two arms separate refiner training from base quality. Primary one-shot models train for 80,000 updates; refiners train for 40,000 updates with the base frozen. All matched arms use batch size four, AdamW with learning rate \(2\times10^{-4}\), weight decay 0.05, \((0.9,0.95)\) momentum coefficients, unit-norm gradient clipping, EMA decay 0.999, unweighted voxel cross-entropy, and FP32. The learning rate warms up linearly over the first 2\% of updates and then remains constant. Primary results use five seeds \(\{0,1,2,3,4\}\). Training-time removal draws uniform input dropout in \([0,0.5]\) and, with probability 0.5, a contiguous forward-FOV sector whose angular fraction is uniform in \([0,0.3]\).

The primary parameter match leaves a training-exposure asymmetry: the iterative system combines an 80,000-update base with 40,000 refiner updates, whereas the wide control receives 80,000 updates. We therefore registered a conservative sensitivity before any corrected wide or refiner checkpoint existed. For seeds \(\{0,1,2\}\), \texttt{wide\_aug\_long\_v2} continues the augmented wide model to 120,000 updates and is reevaluated on the full corruption grid. Because all wide-model parameters remain trainable, this control receives more trainable parameter-updates than the iterative system. It tests whether any iterative advantage survives equal aggregate update count; it does not replace the five-seed primary comparison.

Two parameter-identical mechanism controls use seeds \(\{0,1,2\}\). The first removes logit corruption and random restarts while retaining the same refiner, base, input augmentation, and loss. The second replaces the three-level V-cycle with two levels while retaining the shared smoother and all learned parameters. These controls separate sequential inference from its denoising recipe and test whether the coarsest grid contributes beyond added parameters.

The refiner is trained with discounted supervision over three steps. To teach error correction rather than repeated discrimination, its carried belief is occasionally corrupted by logit noise, confident wrong-class blocks, or free-space erase blocks. Later refinement steps can receive an additional random restart. The three step losses receive relative weights \(0.8^2,0.8,1\); the initial belief is corrupted with probability 0.3 and a later carried belief with probability 0.15. An EMA of the refiner supplies the evaluation weights.

\subsection{Corruption protocol}

For each scene and condition, a deterministic random stream is keyed by the corruption seed and scene index. The resulting input hash must agree across every model and training seed.

\textbf{Contiguous angular removal} masks 10\%, 20\%, 30\%, 40\%, or 50\% of the 180-degree forward angular domain. The sensor is at the forward boundary and lateral center, and the sector start is sampled uniformly over placements that remain inside the forward half-plane. We report both nominal angular severity and the achieved fraction of occupied input returns removed.

\textbf{Uniform thinning} removes each occupied input voxel independently with probability 0.25, 0.50, 0.65, 0.75, or 0.85.

Let r denote normalized metric radial distance from the sensor and s the nominal corruption severity. The two distance-dependent probabilities are

\[
p_{\mathrm{drop}}(r;s)=\min(0.98,0.05+s r),\qquad
p_{\mathrm{add}}(r;s)=0.06s(1-r)^2.
\]

\textbf{Range-dependent attenuation} applies the first function to occupied returns at severities 0.2, 0.4, 0.6, and 0.8. This is explicitly a simulated attenuation stress test, not a calibrated claim about a particular fog sensor.

\textbf{Additive clutter} applies the second function only to originally empty voxels, at severities 0.1, 0.2, 0.3, and 0.5. This axis lies outside the removal-only training augmentation and tests the boundary of observation-model coverage.

Accordingly, ``covered'' in the results refers to exposure to a corruption family, not exact support matching. The higher removal severities explicitly test extrapolation beyond the training range; attenuation and clutter test mechanisms absent from training.

\subsection{Metrics and statistical decisions}

We report occupied-versus-free completion IoU, semantic mIoU over the 19 non-free classes, per-class IoU, and corruption-specific retained/removed/added-region diagnostics. Region metrics are chosen according to target support: observed, official-occluded-but-unobserved, and removed surfaces use semantic mIoU and completion IoU; the predominantly free other-unobserved region uses voxel accuracy and the false-occupied rate. Accuracy evaluation is FP32 at fixed depth \(J=3\). A separate depth sweep evaluates \(J=0,\ldots,8\), where \(J=0\) is the frozen base, and records mean logit-update magnitude and the fraction of valid voxels whose predicted class changes.

The primary unit of model-system replication is the training seed; the primary observational unit is the ordered validation frame. For each contrast, we pair exact scene identifiers and corruption hashes, resample training seeds, and use a circular moving-block bootstrap within each selected seed. The primary block length is 20 frames, with sensitivity at 10 and 40. Moving-block resampling is used because ordinary independent resampling would discard the temporal dependence between neighboring frames \citep{kunsch1989blockbootstrap}. Each interval uses 10,000 replicates. We report the observed mIoU difference, 95\% percentile interval, and the fractions of bootstrap replicates above zero and above the predeclared practical margin of 0.5 mIoU. These fractions are descriptive bootstrap stability summaries, not posterior probabilities.

We call an effect practically positive or negative only when the full interval lies beyond the corresponding margin. We describe practical equivalence as supported only when the full interval lies inside \([-0.5,0.5]\); intervals that cross either a directional or practical boundary are reported as unresolved rather than forced into a win, loss, or tie. The 0.5-point margin is a study-specific decision threshold fixed before protocol-v2 outcomes, intended to keep a precisely estimated but negligible gain from being presented as an iterative-system advantage; it is not proposed as a universal SSC standard.

The sole primary confirmatory contrast is the augmented-base refiner minus the augmented parameter-matched wide model at 40\% angular removal. The other 12 registered contrasts are secondary attribution, transport, or mechanism analyses. Their intervals are not multiplicity-adjusted and are interpreted as effect-size diagnostics rather than 12 additional opportunities to declare the main hypothesis successful.

\subsection{Efficiency protocol}

Batch-1 latency is measured with CUDA events after 30 warm-up passes and 100 synchronized timed passes in each of three fresh processes. We report between-process variation, throughput, peak allocated and reserved GPU memory, trainable and total parameters, device, framework version, precision, and fixed iteration depth. FLOPs are counted for the same input and depth using PyTorch's operator counter, with a multiply-add counted as two FLOPs and the counted operator inventory retained in the artifact. FP32 is the main comparison; AMP is labeled separately.

All reported refinement uses a fixed depth. The implementation exposes a threshold-based stopping heuristic, but it was neither trained nor calibrated under protocol v2; we therefore make no adaptive-compute claim and do not tune that threshold on sequence 08.

\subsection{Outcome-blind qualitative cases}

Qualitative panels are restricted to five evenly spaced validation indices \(\{0,204,408,612,814\}\), fixed before either relevant seed-0 checkpoint or any corrected final evaluation archive existed. The main panel uses index 408 under clean input, 40\% sector removal, 75\% thinning, and clutter 0.5; a supplementary sheet shows all five indices under sector removal. Panels compare the parameter-matched augmented wide model with the augmented-base refiner and show both corrections and regressions. These cases are descriptive: none may be replaced after predictions are visible, and they do not enter a statistical decision.

\section{Results}

The sole primary contrast compares the augmented-base refiner with the parameter-matched augmented wide model at \texttt{occ40}. The refiner gains 0.911 mIoU points, with a 95\% block-20 interval of [0.804, 1.040] that lies entirely above the 0.5-point practical margin. The decision is unchanged with the registered block-length sensitivities: [0.804, 1.030] at length 10 and [0.801, 1.042] at length 40. The refiner costs 4.217M parameters, 214.8 GFLOPs, 10.74 ms, and 0.75 GiB, versus 4.195M, 142.8 GFLOPs, 6.25 ms, and 0.23 GiB for the wide control. Against the three-seed wide model continued to 120,000 updates, the \texttt{occ40} effect remains practically positive at +1.005 with interval [0.859, 1.148]. Thus, the registered result is a compute-for-accuracy gain for the full iterative system under contiguous missing evidence.

All release gates passed before the manuscript was populated. The final inventory contains 703 paired evaluation archives over 815 ordered frames, 12 depth diagnostics, 15 isolated efficiency measurements, and 57 archives for the registered training-exposure control. Checkpoint, evaluator, corruption, scene-pairing, and region-partition audits report no failed item.

The nominal corruption labels are useful shorthand but not exact removal rates. Sector removal at \texttt{occ40} removes 45.53\% of observed returns on average, uniform \texttt{drop75} removes 75.02\%, and range attenuation at \texttt{fog0.8} removes 27.48\%. At \texttt{clutter0.5}, added evidence occupies 0.8073\% of eligible empty locations. Stable valid support is dominated by unobserved space: 9,692,109 voxels are originally observed, 629,973,092 are officially occluded after observed surfaces are removed, and 557,715,822 are other originally unobserved voxels.

\subsection{Main performance and severity boundaries}

\begin{table}[t]
\centering
\caption{Semantic mIoU on the five registered headline conditions. Values are cross-seed mean plus or minus standard deviation; custom models use five seeds.}
\label{tab:main}
\small
\setlength{\tabcolsep}{3pt}
\begin{tabular}{p{0.153\linewidth} p{0.153\linewidth} p{0.153\linewidth} p{0.153\linewidth} p{0.153\linewidth} p{0.153\linewidth}}
\toprule
\textbf{Model system} & \textbf{Clean} & \textbf{Sector} & \textbf{Thinning} & \textbf{Attenuation} & \textbf{Clutter} \\
\midrule
One-shot, clean & 17.61 $\pm$ 0.09 & 8.73 $\pm$ 0.03 & 7.81 $\pm$ 0.21 & 15.26 $\pm$ 0.13 & 1.50 $\pm$ 0.06 \\
One-shot, augmented & 18.31 $\pm$ 0.13 & 9.85 $\pm$ 0.04 & 13.79 $\pm$ 0.14 & 17.27 $\pm$ 0.15 & 1.42 $\pm$ 0.06 \\
Wide one-shot, augmented & 18.45 $\pm$ 0.29 & 9.94 $\pm$ 0.10 & 13.87 $\pm$ 0.25 & 17.40 $\pm$ 0.31 & 1.42 $\pm$ 0.07 \\
Refiner, clean base & 18.69 $\pm$ 0.10 & 10.53 $\pm$ 0.04 & 11.18 $\pm$ 0.18 & 17.37 $\pm$ 0.18 & 1.24 $\pm$ 0.07 \\
Refiner, augmented base & 18.95 $\pm$ 0.14 & 10.85 $\pm$ 0.06 & 14.17 $\pm$ 0.20 & 17.96 $\pm$ 0.15 & 1.04 $\pm$ 0.09 \\
\bottomrule
\end{tabular}
\end{table}

The headline table separates two patterns. Augmentation accounts for most of the recovery under independent thinning: the augmented one-shot model nearly closes the gap to the full iterative system. Under contiguous sector removal, however, the augmented-base refiner retains a visible advantage over both augmented one-shot controls. All systems collapse under additive clutter; the strongest clean and missing-evidence model is not the strongest clutter model.

\begin{figure}[t]
\centering
\includegraphics[width=\linewidth]{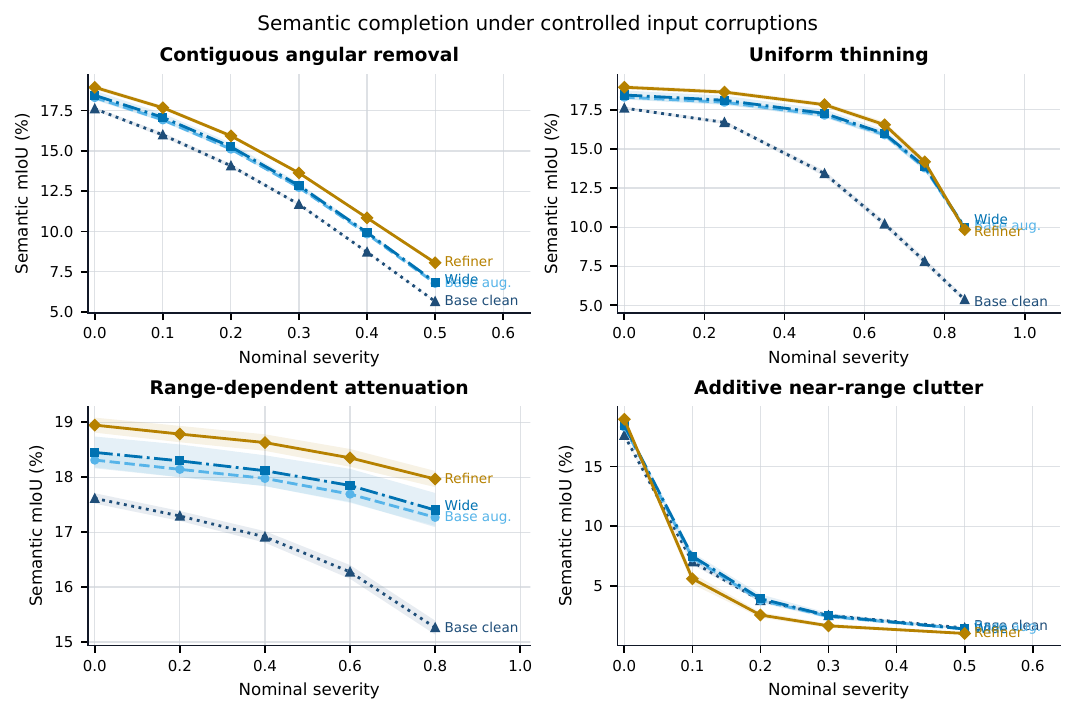}
\caption{Semantic mIoU across the complete registered severity grid. Lines show cross-seed means and bands show one standard deviation. Removal augmentation flattens the thinning and attenuation curves, whereas all model systems deteriorate sharply under additive clutter.}
\label{fig:curves}
\end{figure}

The severity curves confirm that the headline cells are not isolated severity choices. The augmented-base refiner has the highest normalized sector-curve area (14.32 $\pm$ 0.09 mIoU), whereas the wide augmented model reaches 13.56 $\pm$ 0.18. For thinning, the corresponding areas are 17.14 $\pm$ 0.14 and 16.67 $\pm$ 0.25. The ordering reverses under clutter: the refiner's normalized area is 4.25 $\pm$ 0.19, below the wide model's 5.18 $\pm$ 0.22.

\subsection{Iterative-system versus augmentation attribution}

The training-exposure sensitivity also covers clean input and independent thinning. Its clean advantage is +0.762 ([0.527, 0.998]); under \texttt{drop75} the direction remains positive, but the practical magnitude is unresolved at +0.371 ([0.158, 0.565]). These supplementary comparisons do not replace the five-seed primary decision.

\begin{figure}[t]
\centering
\includegraphics[width=\linewidth]{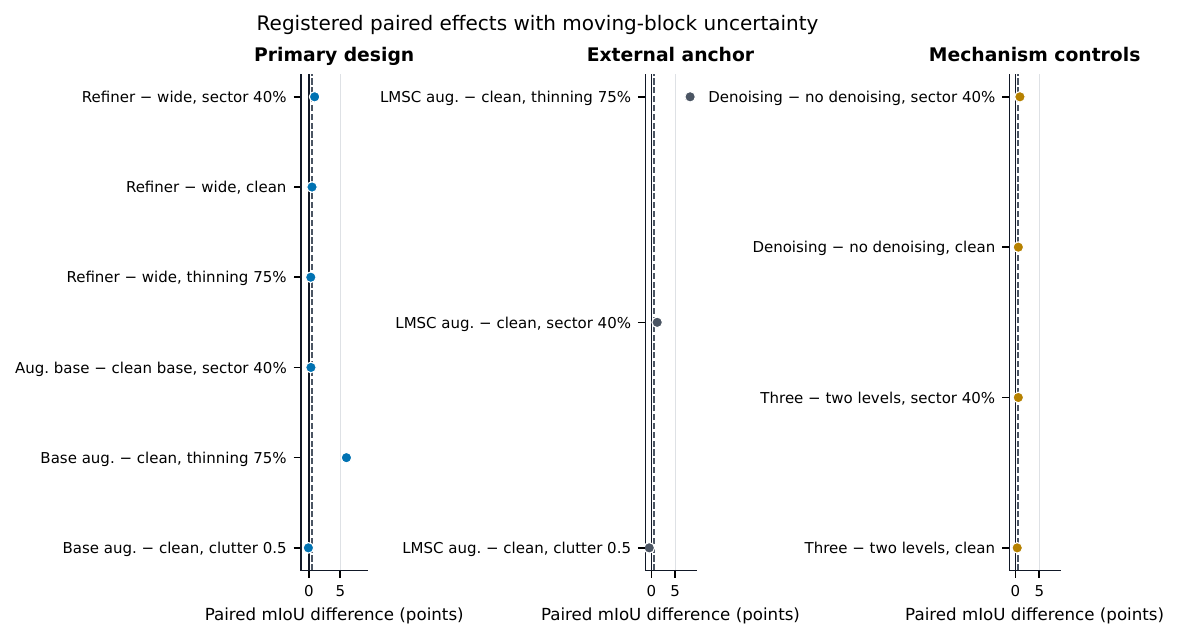}
\caption{Registered paired mIoU effects with 95\% moving-block bootstrap intervals. The vertical solid line marks zero and dotted lines mark the predeclared plus or minus 0.5-mIoU practical margins. The primary comparison is the augmented-base refiner minus the parameter-matched augmented wide model under sector removal.}
\label{fig:effects}
\end{figure}

The secondary contrasts distinguish the main sources of robustness. Adding removal augmentation to the one-shot base yields +5.975 mIoU under \texttt{drop75}, with interval [5.662, 6.140], but slightly lowers clutter performance by 0.082 points, with interval [-0.139, -0.020]. Giving the refiner an augmented rather than clean base adds only +0.325 at \texttt{occ40} ([0.251, 0.405]), a directionally positive but practically small effect.

The mechanism controls show that the primary system cannot be reduced to an arbitrary recurrent wrapper. Relative to the full refiner, omitting the correction curriculum costs 0.931 points at \texttt{occ40} ([0.857, 1.003]), while removing the coarsest recurrent scale costs 0.594 ([0.518, 0.670]). These controls identify dependencies of the complete system; because training recipe and computation are not fully crossed, they do not provide a causal decomposition of iteration alone.

\subsection{Where corrections occur}

The region audit localizes the sector-removal advantage. On originally observed returns removed by \texttt{occ40}, the wide model obtains 5.904 $\pm$ 0.074 semantic mIoU and 28.525 $\pm$ 0.219 completion IoU; the refiner obtains 7.952 $\pm$ 0.165 and 41.990 $\pm$ 1.027. In the stable officially occluded-and-unobserved region, semantic mIoU increases from 8.649 $\pm$ 0.082 to 9.684 $\pm$ 0.053. The gain therefore reaches both the erased surface support and space that was already unobserved, although this diagnostic does not reveal a causal propagation path.

\begin{figure}[t]
\centering
\includegraphics[width=\linewidth]{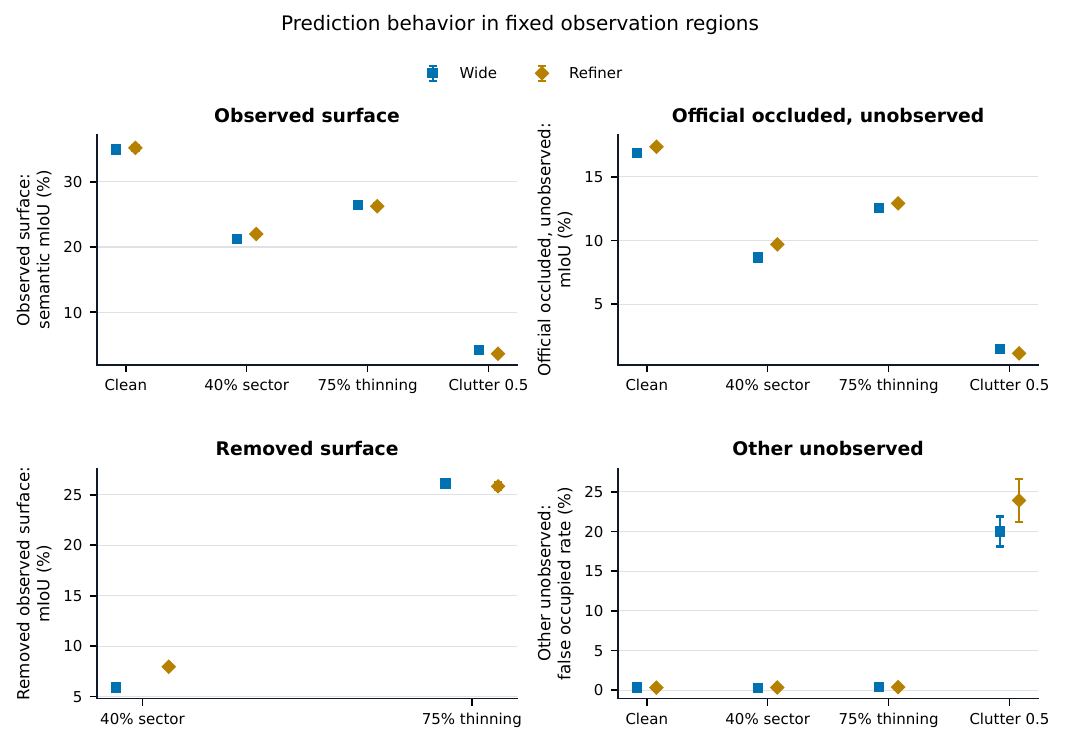}
\caption{Region-specific diagnostics for the augmented wide one-shot model and augmented-base refiner. Regions are fixed from the clean observation before corruption. Removed and added regions isolate locations changed by the corruption.}
\label{fig:regions}
\end{figure}

Clutter exposes the opposite boundary. On artificially added observations, the wide model records 0.858 $\pm$ 0.044 semantic mIoU and a 30.974 $\pm$ 1.437\% false-occupied rate on free targets. The refiner records 0.686 $\pm$ 0.070 mIoU and 33.763 $\pm$ 2.310\% false occupancy. Iteration amplifies rather than repairs some unsupported occupied beliefs.

\subsection{Depth, update stability, and overthinking}

\begin{table}[t]
\centering
\caption{Fixed-depth semantic mIoU for the augmented-base refiner. Depth zero is its frozen one-shot base; values are mean plus or minus standard deviation over three seeds.}
\label{tab:depth}
\small
\setlength{\tabcolsep}{3pt}
\begin{tabular}{p{0.230\linewidth} p{0.230\linewidth} p{0.230\linewidth} p{0.230\linewidth}}
\toprule
\textbf{Condition} & \textbf{Depth zero} & \textbf{Trained depth} & \textbf{Extended depth} \\
\midrule
Clean & 18.30 $\pm$ 0.18 & 18.93 $\pm$ 0.19 & 16.04 $\pm$ 0.69 \\
Sector & 9.87 $\pm$ 0.05 & 10.84 $\pm$ 0.07 & 8.49 $\pm$ 0.16 \\
Thinning & 13.75 $\pm$ 0.17 & 14.07 $\pm$ 0.16 & 11.50 $\pm$ 0.80 \\
Clutter & 1.42 $\pm$ 0.06 & 1.09 $\pm$ 0.08 & 0.95 $\pm$ 0.16 \\
\bottomrule
\end{tabular}
\end{table}

Clean and thinning accuracy peak just before the trained horizon, sector removal peaks at the trained horizon, and clutter is best before refinement. At the extended horizon, every condition is below its earlier maximum. Extra tied computation is therefore not monotonically useful, and the clutter result shows that the learned update can begin from a better one-shot belief and make it worse.

\begin{figure}[t]
\centering
\includegraphics[width=\linewidth]{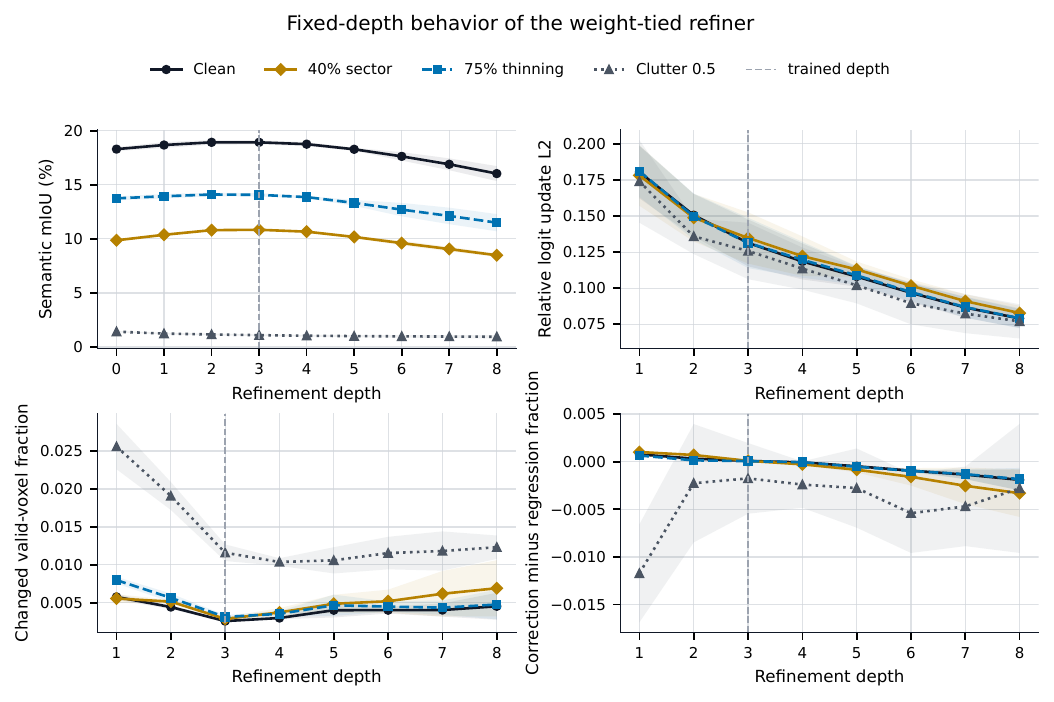}
\caption{Accuracy and update diagnostics beyond the trained three-cycle horizon. Update magnitudes shrink, but corrected and regressed voxel fractions show that smaller updates do not imply improving predictions.}
\label{fig:depth}
\end{figure}

Update magnitude alone would give a misleading stability story. At step 8 the net correction fraction is -0.0019 $\pm$ 0.0011 on clean input and -0.0033 $\pm$ 0.0025 under sector removal: regressions exceed corrections even as the relative update norm declines. Under clutter, the first step is already net negative at -0.0117 $\pm$ 0.0051. The evidence supports an overthinking diagnosis near the trained horizon, not convergence to a contractive fixed point.

\subsection{Accuracy and inference cost}

\begin{table}[t]
\centering
\caption{Accuracy and isolated FP32 batch-one inference cost. Latency is the mean plus or minus between-process standard deviation over three fresh processes.}
\label{tab:cost}
\small
\setlength{\tabcolsep}{3pt}
\begin{tabular}{p{0.153\linewidth} p{0.153\linewidth} p{0.153\linewidth} p{0.153\linewidth} p{0.153\linewidth} p{0.153\linewidth}}
\toprule
\textbf{Model system} & \textbf{Clean mIoU} & \textbf{Parameters (M)} & \textbf{GFLOPs} & \textbf{Latency (ms)} & \textbf{Peak GiB} \\
\midrule
One-shot, augmented & 18.31 $\pm$ 0.13 & 3.683 & 126.0 & 3.18 $\pm$ 0.02 & 0.22 \\
Wide one-shot, augmented & 18.45 $\pm$ 0.29 & 4.195 & 142.8 & 6.25 $\pm$ 0.02 & 0.23 \\
Two-level refiner & 18.57 $\pm$ 0.23 & 4.217 & 213.3 & 9.22 $\pm$ 0.12 & 0.75 \\
Full refiner & 18.95 $\pm$ 0.14 & 4.217 & 214.8 & 10.74 $\pm$ 0.41 & 0.75 \\
\bottomrule
\end{tabular}
\end{table}

\begin{figure}[t]
\centering
\includegraphics[width=\linewidth]{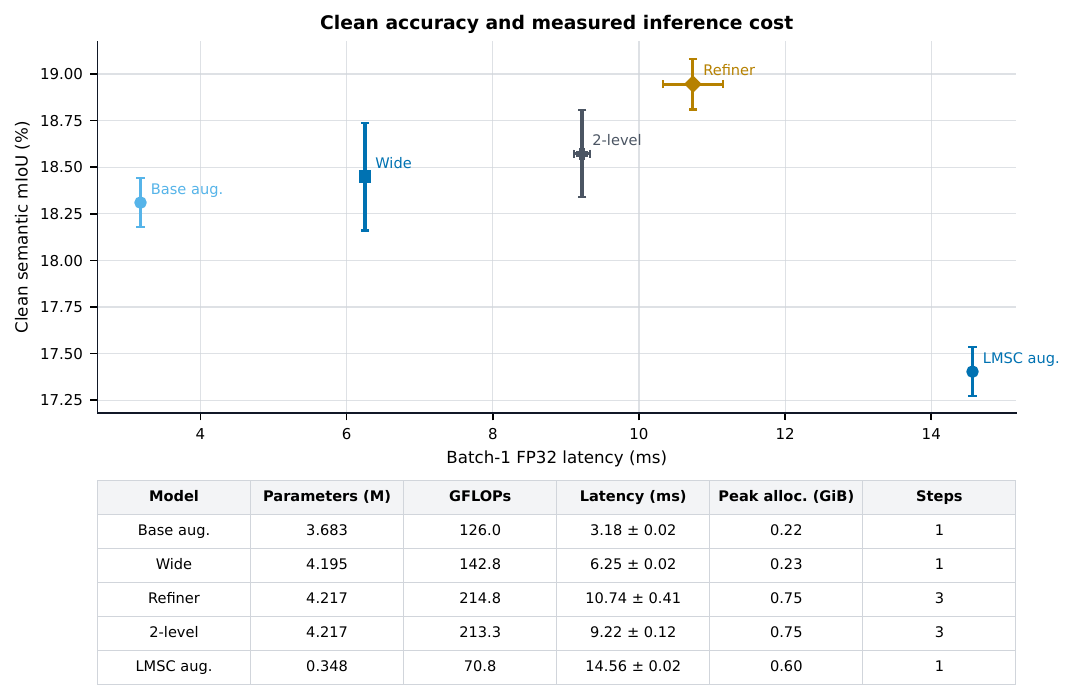}
\caption{Clean accuracy versus measured latency, with marker area indicating peak allocated memory. Parameter matching does not imply compute matching: tied refinement reuses weights but repeats feature processing.}
\label{fig:cost}
\end{figure}

The full refiner is parameter-matched to the wide control but requires substantially more latency and counted FLOPs, with more than triple the peak allocated memory. The two-level diagnostic recovers part of the latency while also giving up accuracy. Any deployment recommendation must therefore treat the primary mIoU gain as a compute-for-accuracy trade rather than a free benefit of weight tying.

\subsection{External-backbone replication of observation coverage}

The recognized LMSCNet-SS topology repeats the observation-coverage pattern. Removal augmentation improves \texttt{drop75} mIoU by 8.153 points ([7.631, 8.705]) and \texttt{occ40} by 1.229 ([1.108, 1.384]); both intervals clear the practical margin. At \texttt{clutter0.5}, however, augmentation changes mIoU by -0.437 ([-0.626, -0.200]), a negative direction whose practical magnitude remains unresolved.

\begin{figure}[t]
\centering
\includegraphics[width=\linewidth]{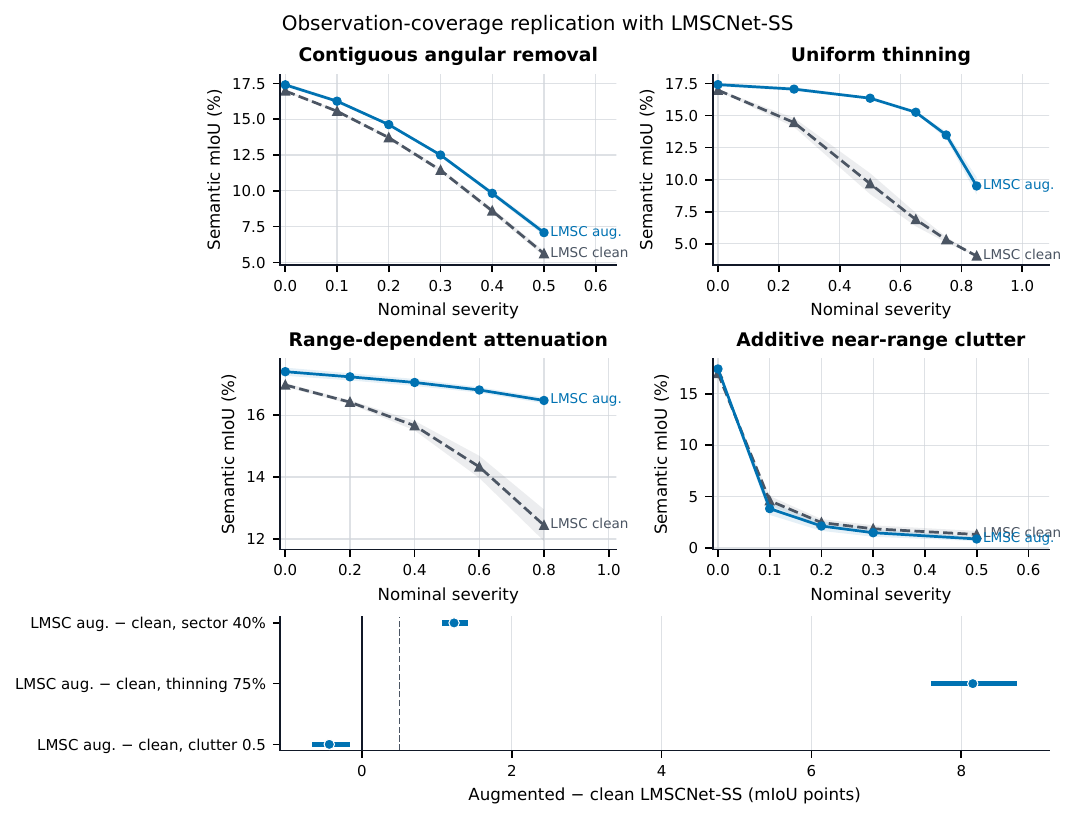}
\caption{Clean and removal-augmented LMSCNet-SS across the registered corruption families. The external topology repeats the benefit of observation-family coverage under missing evidence and the failure to transfer that benefit to additive clutter.}
\label{fig:external}
\end{figure}

This external anchor strengthens the claim that training-distribution coverage drives missing-evidence robustness across more than one compact backbone. It does not test the custom refiner and therefore cannot establish that the iterative-system effect transfers across architectures.

\subsection{Preregistered qualitative cases}

\begin{figure}[t]
\centering
\includegraphics[width=\linewidth]{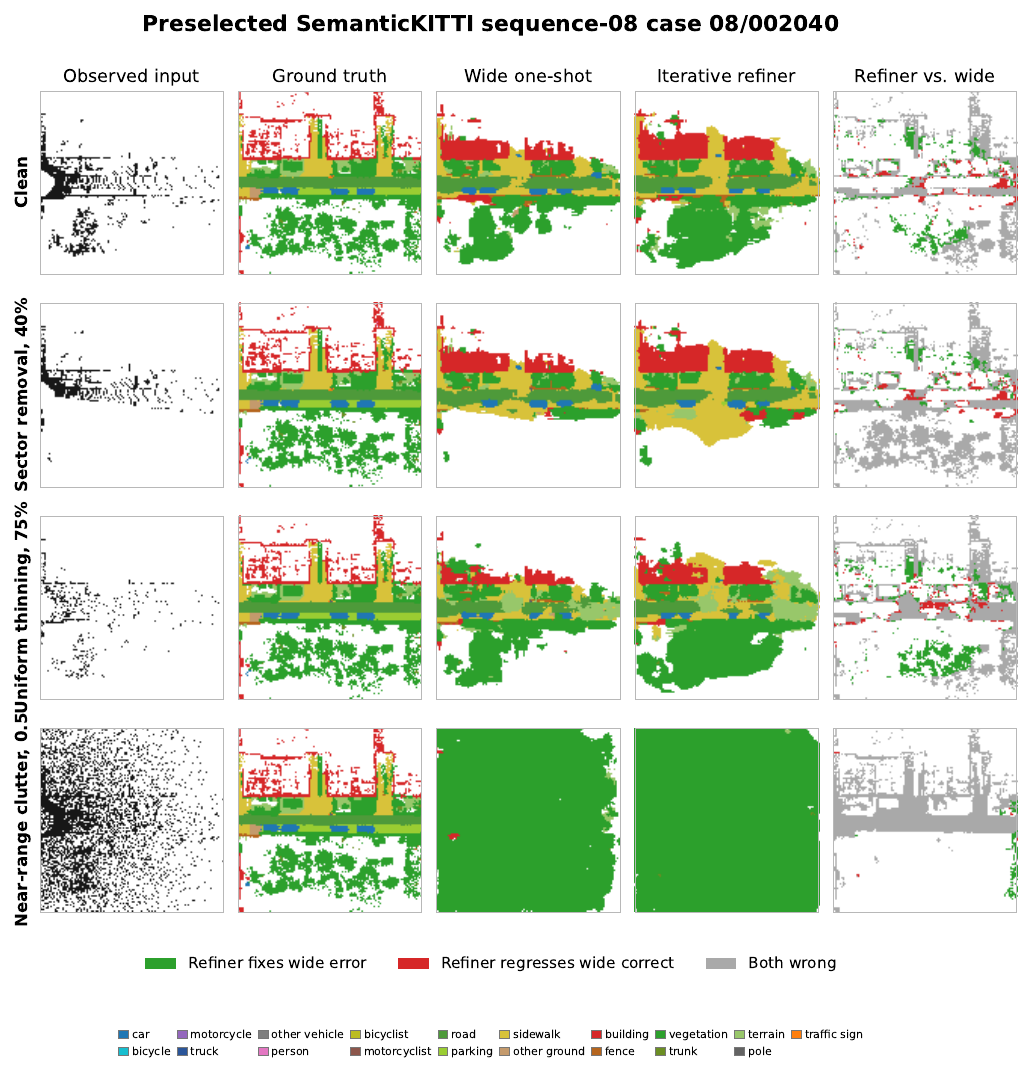}
\caption{Preselected sequence-08 case 08/002040 under the four headline conditions. Green marks voxels where the refiner corrects a wide-model error, red marks regressions, and gray marks shared errors. The case and display rules were fixed before outcomes.}
\label{fig:qualitative}
\end{figure}

The fixed main panel shows corrections and regressions interleaved rather than a uniformly improved object map. Under sector removal and thinning, the refiner fills additional road, vegetation, and building support, but it also changes some wide-model-correct voxels. Under clutter, both systems share a large unsupported occupied prediction and their differences are sparse relative to the common failure. The all-five-scene contact sheet in the appendix preserves the registered sampling frame. These panels illustrate failure geometry only; they are not used to estimate prevalence or define a post hoc subgroup.

\section{Discussion}

The central finding is conditional but clear: fixed-depth iteration helps when the input contains a coherent missing sector, even after parameter count, observation-family augmentation, scene realization, and aggregate update count are controlled. The interval clears the study's practical margin and remains stable across the registered temporal block lengths. Because the primary arm also includes multistep supervision, a correction curriculum, and a three-level hierarchy, the right scientific object is the full iterative system. Calling the result an architecture-only effect would overstate identification.

The augmentation results provide an equally important counterweight. Under independent thinning, teaching the observation family yields a much larger gain than adding iterative inference; the same pattern appears in LMSCNet-SS. This is not evidence that iteration is irrelevant. It says that sequential computation should not receive credit for robustness supplied by the training distribution. The clean-base control makes the same point within the refiner: base quality matters, but it explains only a practically small portion of the sector-removal contrast.

The region and depth diagnostics suggest why the boundary is asymmetric. Refinement changes predictions on erased surface support and in officially unobserved space, consistent with a role for repeated spatial correction. Under clutter, however, the observation itself asserts false occupancy. Input recall then preserves the wrong evidence, and repeated updates can spread it. This interpretation is diagnostic rather than causal, but it aligns the region-specific false-occupancy increase, the negative first update, and the qualitative failure geometry.

More computation is not automatically better. Accuracy peaks near the trained depth and declines when the tied update is unrolled further, despite shrinking update norms. This separates numerical smallness from semantic improvement and argues against presenting the learned dynamics as convergent. It also limits the deployment lesson: the measured sector-removal gain is purchased with substantially higher latency and memory. A wide one-shot model is the more efficient default; the fixed-depth refiner is warranted only when coherent missing evidence is sufficiently important to justify that cost.

The earlier evidence-distance study provides a sharper hypothesis for why contiguous gaps differ from dispersed thinning. Its distance buckets, fixed-area multi-wedge manipulation, and SSCBench--KITTI-360 direction are qualitatively compatible with the corrected geometry-conditioned boundary. They are not pooled with protocol v2, however, because the legacy loader and forward-FOV implementation were corrected later and the old caches lack the final checkpoint/evaluator hash chain. Appendix~\ref{app:exploratory} preserves those analyses as provenance for hypothesis formation, not as confirmatory support.

Taken together, the experiments support a decision rule rather than a universal model ranking. Cover expected missing-evidence families during training first. Add fixed-depth refinement when contiguous evidence gaps remain important and the compute budget permits it. Treat additive clutter as a separate robustness problem requiring rejection, uncertainty, or observation validation rather than simply more iterations.

\section{Deployment Implications}

\begin{enumerate}
\item \textbf{Cover the expected observation family first.} Under independent 75\% thinning, removal augmentation adds 5.975 mIoU to the base model, far more than the 0.300-point refiner--wide difference. The LMSCNet-SS anchor reproduces this training-coverage effect.

\item \textbf{Use widening as the efficient default.} The wide one-shot control runs at 6.25 ms and 0.23 GiB, whereas the full refiner requires 10.74 ms and 0.75 GiB. A fixed-depth refiner is justified when coherent missing regions matter enough for the 0.911-point paired gain to offset that cost.

\item \textbf{Do not extrapolate depth blindly.} Accuracy peaks near the trained three-step horizon and then declines even as update norms shrink. Any adaptive-depth policy should therefore be validated against semantic corrections and regressions, not numerical convergence alone.

\item \textbf{Treat false evidence as a separate failure mode.} Neither removal augmentation nor repeated refinement repairs additive clutter; the refiner can increase false occupancy at inserted returns. Observation validation, rejection, or uncertainty-aware fusion is a more direct target than additional iterations.
\end{enumerate}

\section{Limitations}

SemanticKITTI sequence 08 served as a development set in the legacy phase. Protocol v2 was frozen before corrected retraining, but the final evidence is not equivalent to an untouched hidden test. We label the legacy phase exploratory and the corrected phase confirmatory-within-sequence. A recognized external topology tests whether the observation-coverage result depends on the custom backbone, but it does not replicate the custom iterative-system contrast and does not replace a second dataset or an official test-server submission.

The corruptions are controlled stress tests. Range attenuation and clutter are not calibrated to a specific LiDAR, weather distribution, or failure rate. Their value is mechanistic separation, not a claim of real-world prevalence. Likewise, the three hypotheses organize measurable predictions but do not prove that a learned update is contractive or that a task has a particular intrinsic computational depth.

Finally, the conclusions concern single-frame voxel SSC and compact models. They need not transfer to camera occupancy, sparse high-capacity transformers, multi-frame fusion, tracking, or forecasting, where sequential state updates may be intrinsically required.

The primary refiner--wide comparison does not isolate architecture in the strict causal sense. Multistep supervision and the correction curriculum are parts of the iterative system's training. The no-denoising and two-level controls localize specific dependencies, but they do not form a fully crossed training-recipe design. We therefore interpret the primary difference as a full iterative-system effect and use depth and update diagnostics as supporting evidence rather than causal proof.

Five training seeds support a paired model-system comparison, but they do not characterize the tails of optimization variability. Likewise, a circular moving-block bootstrap over one ordered validation sequence preserves local dependence only approximately; its interval describes the registered seeds and sequence 08 rather than a wider population of routes, sensors, or datasets. We therefore report sensitivity to block lengths 10, 20, and 40 and treat any decision that changes across those lengths as sensitivity-dependent.

The common unweighted objective and FP32 precision were fixed using exploratory legacy observations on the same development sequence. This preserves fairness across the corrected custom-model arms but is not an independent hyperparameter selection. The legacy class-balanced and mixed-precision comparisons used the superseded protocol and are not cited as corrected quantitative evidence.

\section{Reproducibility and Responsible Reporting}

Every run records the protocol version, training arguments, seed, code hash, checkpoint SHA-256, environment, precision, and device. Evaluation archives contain per-scene confusion matrices, sequence/frame ids, corruption hashes, and achieved corruption intensity. A fused evaluator decodes each validation scene once and batches its 19 deterministic corruption views; exact parity with the single-condition reference was verified on two nonadjacent real frames on both local and cloud data copies before the final result directory was opened. The large artifact archive contains the complete run tree, checkpoints, logs, and per-scene sufficient statistics. The lightweight anonymous package contains their hash manifests, the analysis code, portable configurations, compact evidence tables, and every figure source.

An automated release gate loads all 40 final checkpoints and verifies their completion markers, terminal steps, registered arguments, architectures, finite evaluation weights, and unique content hashes. Independent archive audits then verify the frozen result inventory, exact corruption pairing, region-partition identities, and invariant target support before any paper table is compiled.

We do not claim state-of-the-art SSC accuracy. Public checkpoints are labeled official only when provenance and hashes are recorded. Results from the geometry- or label-bugged legacy protocol are retained for audit but excluded from final quantitative claims.

\section{Conclusion}

Test-time iteration is neither a universal robustness mechanism nor a disguised capacity increase in this controlled SSC setting. The complete iterative system provides a practically positive advantage over a parameter-matched wide one-shot model when a contiguous sector of evidence is missing, and that advantage survives the registered training-exposure sensitivity. Independent thinning tells a different story: observation-family augmentation supplies the dominant gain and transfers to LMSCNet-SS without the custom refiner. Additive clutter remains a failure for both strategies.

The deeper lesson is methodological. Claims about iterative inference require matched augmentation, capacity, exact corruptions, training exposure, temporal dependence, inference cost, and behavior beyond the trained depth. With those controls in place, iteration earns a narrow role: it can improve coherent missing-evidence completion when extra compute is acceptable, but it should complement rather than substitute for observation-model coverage and explicit handling of spurious evidence.

\clearpage
\bibliography{merged_references}
\bibliographystyle{tmlr}
\appendix

\section{Pre-run protocol amendment}

See supplementary protocol S1 (\texttt{PROTOCOL\_V2.md}).

\section{Label and geometry audits}

Protocol v2 remaps raw zero to free space, maps nonzero raw ids with learning id zero to ignore, and applies the invalid mask after remapping. A geometry parity check verifies tensor-axis conventions before the final result directory is opened. The corruption audit confirms unique deterministic inputs for all registered scene-condition pairs, and the region audit confirms that stable target support is invariant across every archive.

\section{Full severity and per-class tables}

The compact evidence directory contains the full severity table, normalized curve areas and slopes, per-class scores, paired effects, region diagnostics, depth trajectories, update dynamics, training-exposure controls, and efficiency measurements in machine-readable CSV form. Each table hash is declared in the atomic paper marker and checked by the release gate.

\begin{figure}[t]
\centering
\includegraphics[width=\linewidth]{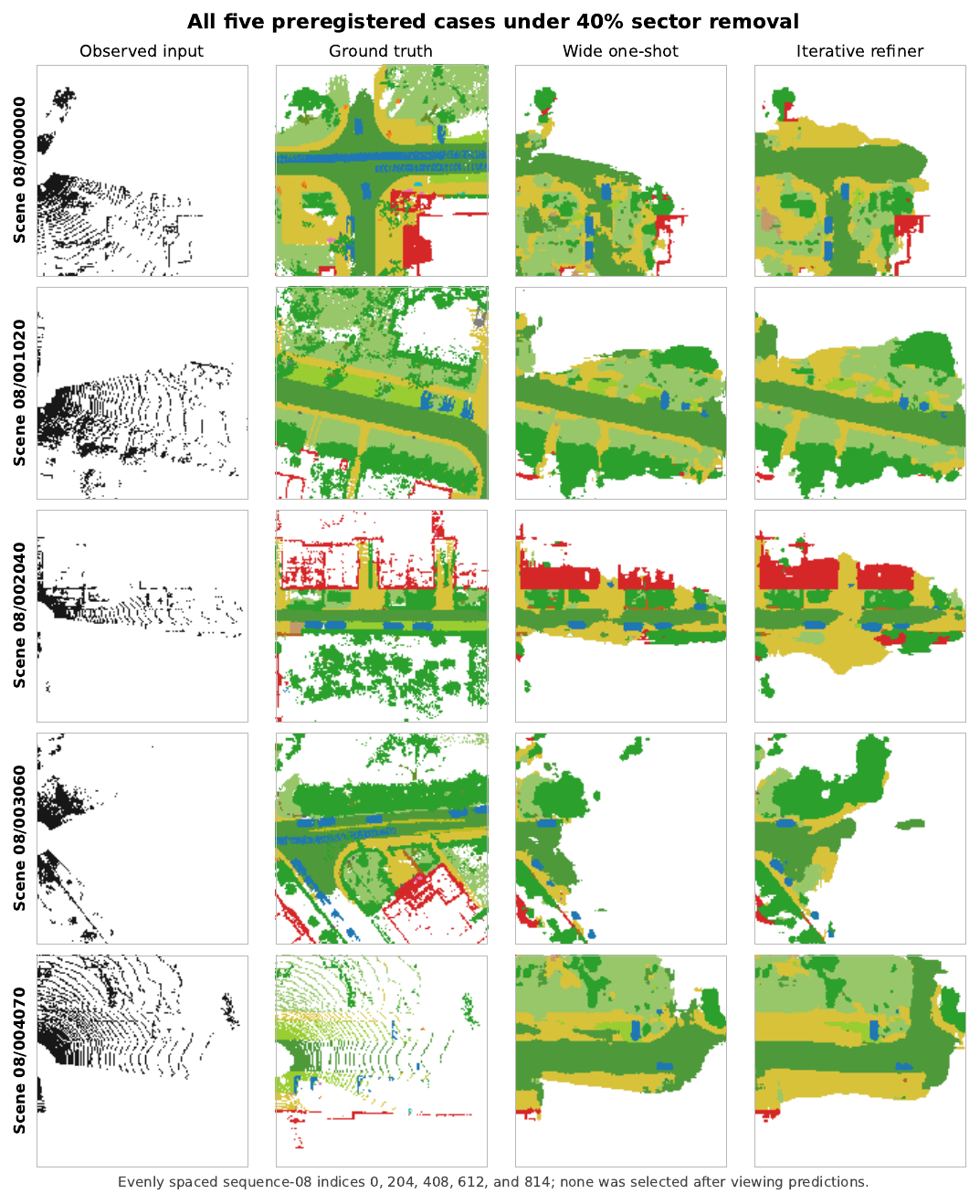}
\caption{All five evenly spaced preregistered sequence-08 cases under sector removal. No scene was replaced after predictions were visible.}
\label{fig:qualitative-supp}
\end{figure}

\section{Objective and precision selection provenance}

Unweighted cross-entropy and FP32 were fixed from exploratory legacy work before corrected protocol-v2 retraining. The corrected arms share those choices for fairness; legacy class-balanced, mixed-precision, geometry-bugged, and label-bugged numbers are excluded from the manuscript's quantitative evidence.

\section{Reproducibility checklist}

\begin{itemize}
\item The dataset split, labels, voxel geometry, corruption seeds, and achieved
intensities are recorded.

\item Every model records its arguments, seed, terminal step, checkpoint hash, and
code hash.

\item Paired archives retain scene IDs, corruption hashes, sufficient statistics,
and fixed region masks.

\item The statistical record fixes the primary contrast, practical margin,
bootstrap count, block length, and sensitivities.

\item Accuracy, parameter and FLOP cost, latency, memory, depth behavior, and
qualitative failures are released together.

\item The anonymous release contains code, configurations, compact evidence,
citations, source, and hashes; the archive adds checkpoints and logs.
\end{itemize}

\clearpage
\section{Exploratory evidence-distance and cross-dataset study}
\label{app:exploratory}

An earlier draft asked whether distance to the nearest surviving observation could explain the geometry-conditioned result. It reported distance buckets, a fixed-area multi-wedge manipulation, per-step update fronts, and a three-seed SSCBench--KITTI-360 replication. These analyses remain useful for formulating a sharper hypothesis and designing a corrected experiment.

They are \emph{not claim-bearing evidence in this paper}. They were produced with the superseded protocol-v1 loader and/or unversioned result caches, before the official label-semantic and forward-FOV geometry audits. The legacy loader treated some nonzero raw labels mapped to learning id zero as free rather than ignored, and the legacy single-sector implementation sampled a 360-degree domain although the SSC grid covers the forward 180-degree field of view. The caches also lack the final protocol version, checkpoint SHA-256 values, evaluator hash, and per-scene sufficient-statistics chain required by protocol v2. We therefore do not combine their numerical estimates with the corrected five-seed intervals and do not claim corrected cross-dataset replication.

\begin{figure}[h]
\centering
\begin{minipage}[t]{.49\linewidth}
\includegraphics[width=\linewidth]{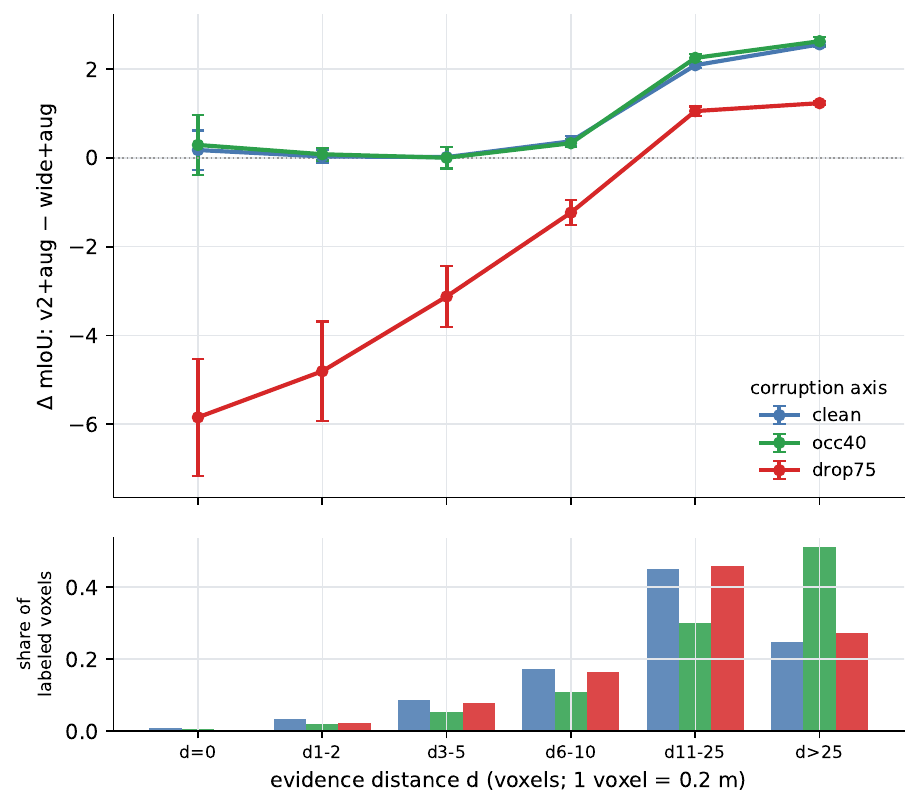}\\
\small (a) Distance-stratified paired differences
\end{minipage}\hfill
\begin{minipage}[t]{.49\linewidth}
\includegraphics[width=\linewidth]{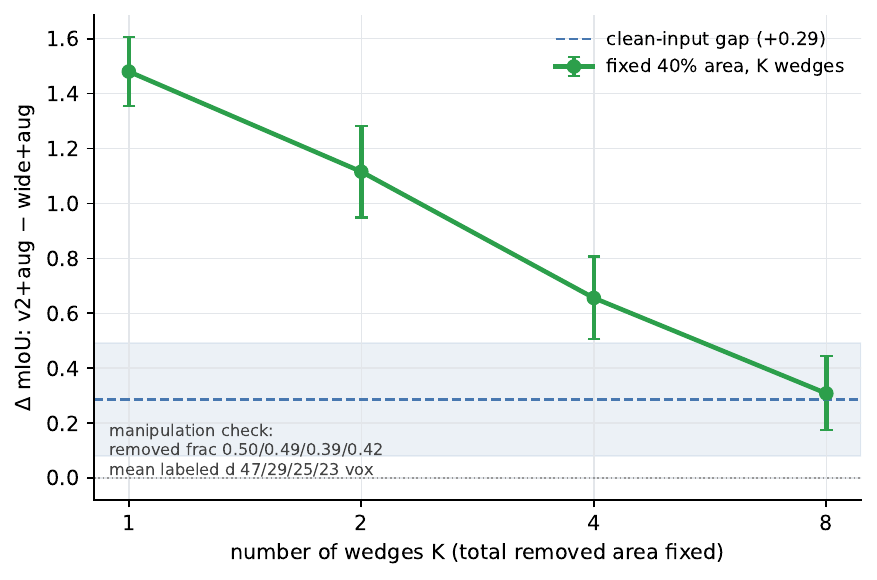}\\
\small (b) Fixed-area multi-wedge manipulation
\end{minipage}
\caption{Protocol-v1 exploratory analyses retained to document hypothesis formation. Values are excluded from the paper's confirmatory claims.}
\label{fig:exploratory-distance}
\end{figure}

\begin{figure}[!t]
\centering
\includegraphics[width=\linewidth]{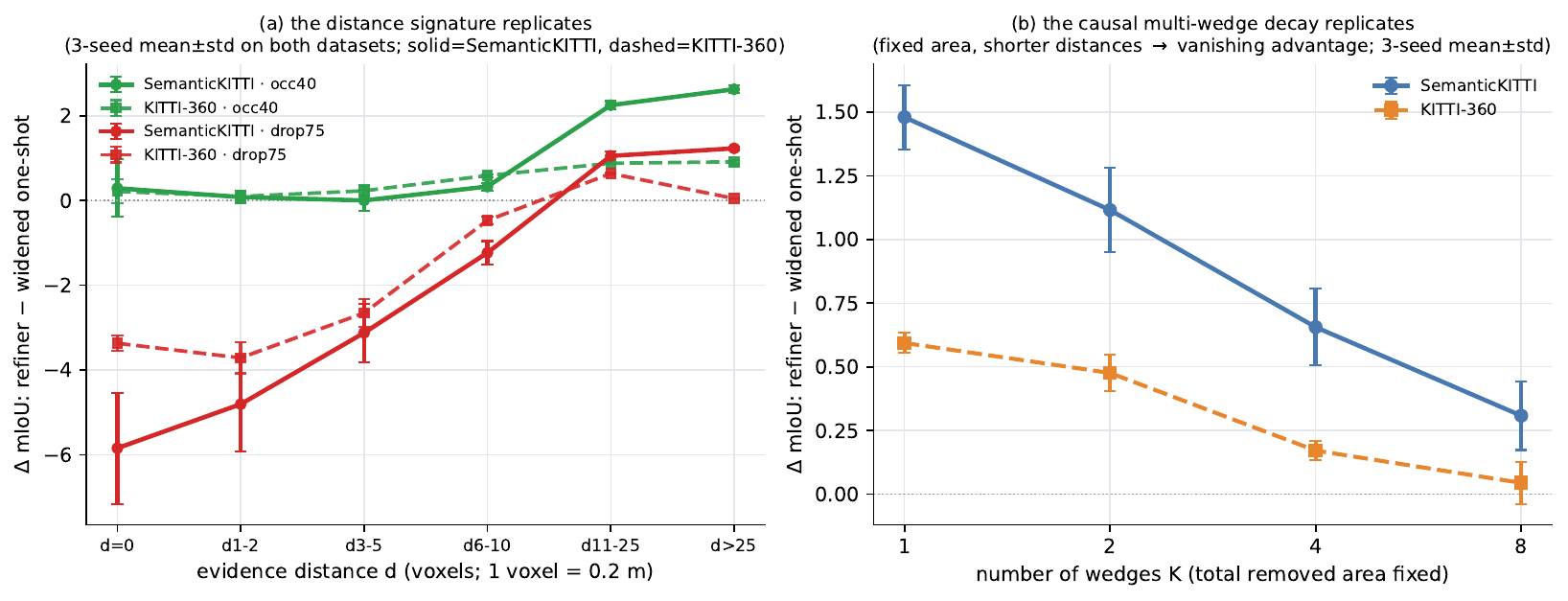}
\caption{Protocol-v1 exploratory SSCBench--KITTI-360 analysis. It motivates a future corrected replication but is not presented as validated transfer.}
\label{fig:exploratory-kitti360}
\end{figure}

A claim-bearing test of the distance hypothesis would rerun the single-wedge, fixed-area multi-wedge, distance-stratified, and KITTI-360 evaluations under protocol v2 with frozen checkpoints, corrected label semantics and forward geometry, achieved-intensity audits, exact scene-level pairing, and the same hierarchical bootstrap used in the main study.

\end{document}